\documentclass[11pt]{article}

\usepackage[final]{acl}
\usepackage{times}
\usepackage{latexsym}
\usepackage[T1]{fontenc}
\usepackage[utf8]{inputenc}
\usepackage{microtype}
\usepackage{inconsolata}
\usepackage{graphicx}
\usepackage{booktabs}
\usepackage{multirow}
\usepackage{amsmath}
\usepackage{xcolor}
\usepackage{float}
\usepackage{tikz}
\usetikzlibrary{arrows.meta,positioning}

\title{PSK at WMT 2026 MIST: Task-Specialized QLoRA Adapters for Multilingual Summarization and Question Answering}

\author{
  Srikar Kashyap Pulipaka \\
  Independent Researcher \\
  \texttt{srikar.kashyap@gmail.com}
}

\begin{document}
\maketitle

\begin{abstract}
We describe the PSK submission to the WMT 2026 Multilingual Instruction Shared Task. Our system uses the 3.35B-parameter Tiny Aya Global model with three QLoRA adapters, one for each task. The adapters are trained on multilingual document--summary pairs, passage-based question answering, and filtered standalone question answering. The summarization data also includes scientific papers with their author-written abstracts. On our held-out split, the context and summarization adapters perform better than our multitask adapter, which was trained only on data supplied by the organizers. Results for open QA are mixed and vary with answer length and evaluation method. We therefore submit three systems with the same context and summarization adapters but different open-QA adapters.
\end{abstract}

\section{Introduction}

The WMT 2026 Multilingual Instruction Shared Task (MIST) evaluates models under a 10B-parameter limit on three tasks: context-based question answering, summarization from a document in language $X$ to language $Y$, and open-ended generation \cite{wmt26mist}. The test set covers 24 languages and contains both same-language and cross-lingual generation. MIST follows the broader multilingual evaluation effort introduced in WMT 2025, which found substantial variation across tasks and languages and also showed that automatic metrics are imperfect for open generation \cite{kocmi-etal-2025-findings-wmt25}.

We use one multilingual backbone with three task-specific adapters. Tiny Aya Global \cite{salamanca-etal-2026-tinyaya} is first adapted to the provided data and then continued separately for each task. This design preserves a single 3.35B backbone while allowing the supervision and decoding policy to follow the output structure of each task.

\section{Task and System Overview}

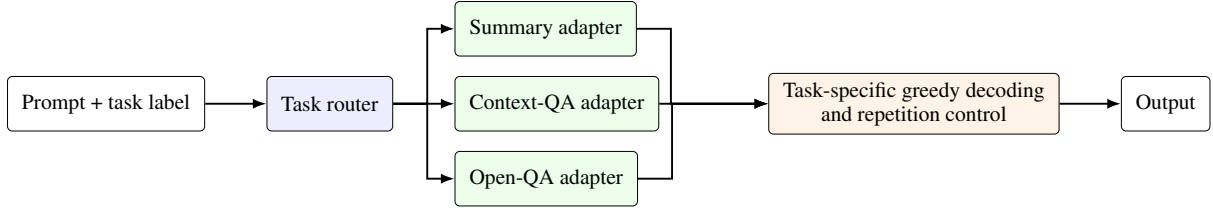
\begin{figure*}[t]
\centering
\resizebox{\textwidth}{!}{%
\begin{tikzpicture}[
  node distance=7mm and 9mm,
  box/.style={draw, rounded corners=2pt, minimum height=8mm, align=center, inner xsep=6pt, font=\small},
  route/.style={box, fill=blue!7},
  adapter/.style={box, fill=green!8},
  decode/.style={box, fill=orange!10},
  arrow/.style={-{Latex[length=2mm]}, thick}
]
\node[box] (input) {Prompt + task label};
\node[route, right=of input] (router) {Task router};
\node[adapter, right=of router, yshift=11mm] (sum) {Summary adapter};
\node[adapter, right=of router] (ctx) {Context-QA adapter};
\node[adapter, right=of router, yshift=-11mm] (open) {Open-QA adapter};
\node[decode, right=16mm of ctx] (decode) {Task-specific greedy decoding\\and repetition control};
\node[box, right=of decode] (output) {Output};
\draw[arrow] (input) -- (router);
\draw[arrow] (router.east) -- ++(5mm,0) |- (sum.west);
\draw[arrow] (router) -- (ctx);
\draw[arrow] (router.east) -- ++(5mm,0) |- (open.west);
\draw[arrow] (sum.east) -- ++(5mm,0) |- (decode.west);
\draw[arrow] (ctx) -- (decode);
\draw[arrow] (open.east) -- ++(5mm,0) |- (decode.west);
\draw[arrow] (decode) -- (output);
\end{tikzpicture}
}
\caption{The three routes share one frozen Tiny Aya Global backbone. The known task label selects an adapter and its decoding policy; language does not change the route.}
\label{fig:pipeline}
\end{figure*}

MIST provides a task label at inference time. We use it to select one of three LoRA adapters and the corresponding decoding configuration (Figure~\ref{fig:pipeline}); there is no language-specific routing.

\subsection{Base Model and Initial Adaptation}

We use \texttt{CohereLabs/tiny-aya-global}, a 3.35B-parameter model covering 70 languages \cite{salamanca-etal-2026-tinyaya}. Its 8K context window supports the longer task inputs. An initial multitask adapter is trained on 20,293 provided examples using a deterministic 85/15 split; the resulting 3,626 held-out examples are reused for model selection. The task-specific adapters are initialized from this model.

\section{Training Data}

Table~\ref{tab:data} summarizes the task-specific training data. Validation examples and exact test prompts are excluded from training.

\begin{table*}[t]
\centering
\small
\begin{tabular}{lrlp{8.4cm}}
\toprule
\textbf{Adapter} & \textbf{Rows} & \textbf{Context} & \textbf{Main sources and construction} \\
\midrule
Multitask adapter & 20,293 & 8,192 & Provided MIST sample training split across all three tasks \\
Summarization & 12,000 & 8,192 & 6,000 general examples from the provided data, CrossSum, WikiLingua, UPDESH, and targeted language pools; 6,000 ACL papers paired with author abstracts \\
Context QA & 8,512 & 4,096 & Belebele, answerable TyDi QA, MLQA, MCIF, UPDESH cultural multihop data, and small Czech/Yoruba Aya pools \\
Open QA & 6,400 & 2,048 & Provided open-QA examples plus filtered Aya and WMT25 MIST examples; includes Hindi questions and answers translated into Bhojpuri \\
\bottomrule
\end{tabular}
\caption{Supervised data used by the selected task adapters. Each is initialized from the multitask adapter.}
\label{tab:data}
\end{table*}

\subsection{Summarization}

The 12,000-example mixture is divided evenly between general and scientific text. General examples combine provided data with CrossSum \cite{bhattacharjee-etal-2023-crosssum}, WikiLingua \cite{ladhak-etal-2020-wikilingua}, and UPDESH \cite{chitale-etal-2026-updesh}. The scientific portion uses full ACL Anthology papers \cite{bird-etal-2008-acl} with author-written abstracts. No Language Left Behind (NLLB) \cite{nllb-team-2024-scaling} supplies non-English targets, which are filtered for language, length, repetition, and semantic consistency. We remove test overlaps and placeholder-heavy papers. We also screen the provided data for likely misalignment by comparing each summary with chunks of its source document and discarding a pair only when both its best LaBSE similarity and best chrF score fall below fixed thresholds.

\subsection{Context Question Answering}

The context-QA training set combines Belebele \cite{bandarkar-etal-2024-belebele}, answerable TyDi QA \cite{clark-etal-2020-tydi}, MLQA \cite{lewis-etal-2020-mlqa}, MCIF \cite{papi-etal-2026-mcif}, UPDESH, and small Czech and Yoruba Aya subsets. MLQA supplies parallel and cross-lingual examples.

\subsection{Open-Ended Question Answering}

Open QA combines provided examples, WMT25 MIST questions, and the Aya Dataset and Collection \cite{singh-etal-2024-aya-dataset}. We retain short factual and explanatory QA while excluding passage-dependent, multiple-choice, creative-writing, code, translation, and continuation tasks. We also translate a small set of Hindi questions and answers into Bhojpuri with NLLB.

\section{Training and Inference}

We use QLoRA \cite{dettmers-etal-2023-qlora} with 4-bit NF4 quantization and BF16 computation. LoRA \cite{hu-etal-2022-lora} is applied to the attention and feed-forward projections with rank 16, alpha 32, and dropout 0.05. Each adapter is trained for one epoch with effective batch size 16; learning rates are $1\times10^{-4}$ for summarization, $8\times10^{-5}$ for open QA, and $2\times10^{-4}$ for context QA.

Inference uses greedy decoding without sampling. After observing repetitive outputs on the development set, we add repetition penalties of 1.05 for open QA and 1.03 for summarization. We also block repeated 4-grams in summarization outputs.

\section{Development Results}
\label{sec:results}

We report exact match (EM), chrF \cite{popovic-2015-chrf}, ROUGE-L \cite{lin-2004-rouge}, and LaBSE cosine similarity \cite{feng-etal-2022-language}. For open QA, we use chrF, ROUGE-L, and LaBSE together with manual inspection of validation outputs.

\subsection{Task-Level Results}

Table~\ref{tab:main-results} reports development scores for the evaluated adapters. The 8.5k context mixture and the 12k summary mixture are selected for submission.

\begin{table*}[t]
\centering
\small
\begin{tabular}{llrrrrr}
\toprule
\textbf{Task} & \textbf{Training variant} & \textbf{$n$} & \textbf{EM} & \textbf{chrF} & \textbf{ROUGE-L} & \textbf{LaBSE} \\
\midrule
\multirow{2}{*}{Summarization} & Multitask adapter & 1,651 & 0.00 & 22.97 & 0.1571 & 0.6681 \\
 & \textbf{12k general/scientific mix} & 1,651 & 0.00 & \textbf{26.81} & \textbf{0.1780} & \textbf{0.6919} \\
\midrule
\multirow{3}{*}{Context QA} & Multitask adapter & 1,517 & 66.25 & 77.07 & 0.6834 & 0.8813 \\
 & \textbf{8.5k context mixture} & 1,517 & \textbf{68.89} & \textbf{78.58} & 0.6959 & 0.8897 \\
 & +4.8k targeted continuation & 1,517 & 68.36 & 78.36 & \textbf{0.6966} & \textbf{0.8906} \\
\midrule
\multirow{3}{*}{Open QA} & Multitask adapter & 794 & 3.90 & 28.34 & 0.2287 & 0.6524 \\
 & Long-form QA & 794 & 4.03 & 27.96 & 0.2307 & 0.6600 \\
 & \textbf{Best-score QA} & 794 & \textbf{4.16} & \textbf{28.41} & \textbf{0.2329} & \textbf{0.6638} \\
\bottomrule
\end{tabular}
\caption{Internal model-selection results on identical held-out rows. EM is a percentage and LaBSE is cosine similarity.}
\label{tab:main-results}
\end{table*}

The targeted context continuation is marginally higher on ROUGE-L and LaBSE, whereas the 8.5k context mixture is higher on EM and chrF. We therefore select the latter.

\subsection{Scientific Summarization}

Table~\ref{tab:science} reports results on a 336-example scientific validation set. Gold targets are author abstracts; non-English targets are translated and quality filtered. The 12k mixture is strongest across all three metrics.

\begin{table}[t]
\centering
\small
\begin{tabular}{lrrr}
\toprule
\textbf{Model} & \textbf{chrF} & \textbf{ROUGE-L} & \textbf{LaBSE} \\
\midrule
Multitask adapter & 13.27 & 0.0742 & 0.6556 \\
8k summary mix & 25.28 & 0.1812 & 0.6959 \\
\textbf{12k summary mix} & \textbf{29.76} & \textbf{0.2082} & \textbf{0.7535} \\
\bottomrule
\end{tabular}
\caption{Results on the scientific summarization validation set ($n=336$). The summary mixtures are evenly divided between general and scientific data.}
\label{tab:science}
\end{table}

Blocking repeated 4-grams removes the observed summary loops.

\subsection{Open QA and Submitted Systems}

Best-score QA has the strongest aggregate automatic scores. Long-form QA is stronger on a manually identified set of long-form prompts and reaches the generation limit less often. Our main submission uses Long-form QA. A second submission uses Best-score QA. Table~\ref{tab:systems} summarizes the three submitted routes.

\begin{table}[t]
\centering
\small
\begin{tabular}{llll}
\toprule
\textbf{System} & \textbf{Summary} & \textbf{Context} & \textbf{Open} \\
\midrule
PSK-primary & 12k mix & 8.5k mix & Long-form \\
PSK-variant\_A & 12k mix & 8.5k mix & Best-score \\
PSK-variant\_B & 12k mix & 8.5k mix & Multitask \\
\bottomrule
\end{tabular}
\caption{Submitted routes. The open-QA variants are named for their selection criterion.}
\label{tab:systems}
\end{table}

\section{Analysis}

Separate adapters improve context QA and summarization on our development set. Scientific summarization also improves after adding papers and their abstracts to the training data. Open QA has no clear winner. Best-score QA performs better on automatic metrics, while Long-form QA works better on the longer questions we checked manually and is less likely to hit the output limit.

The shared summarization route improves chrF over the multitask adapter in all 28 represented languages. The shared context-QA route improves exact match in 19 of 25 languages, ties in four, and declines in two. Open QA remains mixed: the primary route improves chrF in 9 of 19 languages, and variant A improves it in 10 of 19. Appendix~\ref{sec:language-deltas} gives the complete language-level changes and cell sizes.

\section{Conclusion}

We present a routed Tiny Aya Global system with one QLoRA adapter per MIST task. Development results select the 8.5k context mixture and the 12k summary mixture, while the three submissions vary the less stable open-QA route. Official evaluation will provide the appropriate shared-task comparison.

\section*{Limitations}

Reference metrics do not directly measure factuality and are particularly limited for open QA. Translated scientific targets may retain artifacts. Results use one backbone and one development split. The router also assumes that the task label is available.

\section*{Ethics Statement}

Public and challenge-provided datasets are used under their respective terms. Machine translation may reproduce source and model biases, and open-ended outputs may contain incorrect or harmful claims. The system should not be treated as a factual authority without verification.

\bibliography{references}

\clearpage
\appendix
\raggedbottom
\section{Language-Level Changes}
\label{sec:language-deltas}

Tables~\ref{tab:language-shared-deltas} and~\ref{tab:language-open-deltas} report changes from the multitask adapter on identical held-out examples. All submitted systems share the summary and context routes. The primary system uses Long-form QA, variant A uses Best-score QA, and variant B retains the multitask open-QA route, for which every open-QA change is zero. Context-QA changes are exact-match percentage points; summary and open-QA changes are chrF points. These are descriptive results, and some language cells are small.

\begin{table}[H]
\centering
\small
\setlength{\tabcolsep}{3pt}
\begin{tabular}{@{}lrrrr@{}}
\toprule
\textbf{Language} & \textbf{$n_S$} & \textbf{$\Delta_S$} & \textbf{$n_C$} & \textbf{$\Delta_C$} \\
\midrule
Arabic & 85 & +4.39 & 90 & +4.44 \\
Bengali & 66 & +8.18 & 20 & +10.00 \\
Bhojpuri & 14 & +26.89 & -- & -- \\
Czech & 14 & +12.80 & -- & -- \\
Central Kurdish & 14 & +9.23 & 45 & -2.22 \\
German & 34 & +5.97 & 93 & +3.23 \\
English & 70 & +0.87 & 93 & +3.23 \\
Finnish & 14 & +6.59 & 45 & +2.22 \\
French & 68 & +1.11 & 45 & +0.00 \\
Haitian Creole & 34 & +1.86 & 45 & +2.22 \\
Hindi & 85 & +5.40 & 45 & +2.22 \\
Indonesian & 85 & +3.26 & 90 & +4.44 \\
Italian & 34 & +3.83 & 93 & +2.15 \\
Japanese & 82 & +3.02 & 90 & +4.44 \\
Korean & 79 & +2.03 & 90 & -1.11 \\
Marathi & 79 & +5.53 & 45 & +0.00 \\
Persian & 69 & +5.59 & 45 & +6.67 \\
Portuguese & 84 & +0.09 & 45 & +2.22 \\
Russian & 85 & +5.10 & 90 & +2.22 \\
Slovak & 34 & +8.72 & 45 & +2.22 \\
Spanish & 85 & +4.17 & 45 & +4.44 \\
Swahili & 53 & +1.88 & 45 & +4.44 \\
Telugu & 53 & +0.82 & 45 & +0.00 \\
Thai & 48 & +2.30 & 45 & +2.22 \\
Turkish & 84 & +2.45 & 45 & +8.89 \\
Vietnamese & 84 & +4.07 & 45 & +2.22 \\
Yoruba & 47 & +2.60 & -- & -- \\
Chinese & 68 & +1.93 & 93 & +0.00 \\
\bottomrule
\end{tabular}
\caption{Changes for the shared routes. $S$ denotes summarization ($\Delta$chrF), and $C$ denotes context QA ($\Delta$EM in percentage points).}
\label{tab:language-shared-deltas}
\end{table}

\begin{table}[H]
\centering
\small
\setlength{\tabcolsep}{4pt}
\begin{tabular}{@{}lrrr@{}}
\toprule
\textbf{Language} & \textbf{$n$} & \textbf{Primary $\Delta$chrF} & \textbf{Var. A $\Delta$chrF} \\
\midrule
Arabic & 52 & +0.22 & -0.16 \\
Bengali & 11 & +0.15 & -0.72 \\
Czech & 11 & +1.53 & +1.62 \\
German & 43 & -2.02 & -1.34 \\
English & 11 & -0.18 & +1.82 \\
French & 45 & -1.89 & -0.79 \\
Hindi & 52 & -1.44 & +1.75 \\
Indonesian & 53 & +0.86 & +1.50 \\
Italian & 45 & +2.94 & +2.16 \\
Japanese & 52 & -1.25 & -0.84 \\
Korean & 45 & -0.43 & +0.02 \\
Marathi & 45 & -4.24 & -2.62 \\
Portuguese & 45 & -0.70 & -0.68 \\
Russian & 52 & -1.65 & -1.11 \\
Spanish & 45 & -2.42 & -1.57 \\
Turkish & 45 & +0.21 & +1.33 \\
Vietnamese & 45 & +1.62 & +1.00 \\
Yoruba & 45 & +1.87 & +1.57 \\
Chinese & 52 & +1.38 & +0.03 \\
\bottomrule
\end{tabular}
\caption{Open-QA chrF changes. The primary system uses Long-form QA; variant A uses Best-score QA. Variant B is the multitask reference and has zero change.}
\label{tab:language-open-deltas}
\end{table}

\end{document}